\documentclass{article}

\usepackage[numbers]{natbib} 
\usepackage{arxiv}
\usepackage{amsmath}
\usepackage[utf8]{inputenc} % allow utf-8 input
\usepackage[T1]{fontenc}    % use 8-bit T1 fonts
\usepackage{hyperref}       % hyperlinks
\usepackage{url}            % simple URL typesetting
\usepackage{booktabs}       % professional-quality tables
\usepackage{amsfonts}       % blackboard math symbols
\usepackage{nicefrac}       % compact symbols for 1/2, etc.
\usepackage{microtype}      % microtypography
\usepackage{lipsum}		% Can be removed after putting your text content
\usepackage{graphicx}
\usepackage{natbib}
\usepackage{doi}

\title{Estimation of control area in badminton doubles with pose information from top and back view drone videos}

\date{}	% Here you can change the date presented in the paper title
\author{ \href{https://orcid.org/0000-0002-3067-7341}{\includegraphics[scale=0.06]{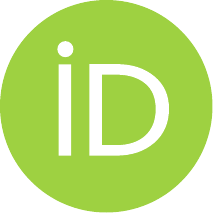}\hspace{1mm} Ning Ding}\thanks{Corresponding author} \\
	Graduate School of Informatics\\
	Nagoya University\\
	  Aichi, Japan \\
	\texttt{ding.ning@g.sp.m.is.nagoya-u.ac.jp} \\
        \And
        {\hspace{1mm}Kazuya Takeda} \\
	Graduate School of Informatics\\
	Nagoya University\\
	Aichi, Japan \\
	\texttt{kazuya.takeda@nagoya-u.jp}
        \And
	{\hspace{1mm}Wenhui Jin} \\
	Department of Physical Education\\
	Wuhu Institute of Technology\\
	Anhui, China \\
	\texttt{jinwenhui@whit.edu.cn} 
	\And
	{\hspace{1mm}Yingjiu Bei} \\
	School of Sports Science\\
	  Anhui Normal University\\
	Anhui, China \\
	\texttt{beibei@mail.ahnu.edu.cn} 
        \And
	\href{https://orcid.org/0000-0001-5487-4297}{\includegraphics[scale=0.06]{orcid.pdf}\hspace{1mm}Keisuke Fujii} \\
	Graduate School of Informatics\\
	Nagoya University\\
	Aichi, Japan \\
	\texttt{fujii@i.nagoya-u.ac.jp} \\
	}

\renewcommand{\headeright}{}
\renewcommand{\undertitle}{}
\hypersetup{
pdftitle={A template for the arxiv style},
pdfsubject={q-bio.NC, q-bio.QM},
pdfauthor={Ning Ding, Kazuya Takeda, Wenhui Jin, Yingjiu Bei, Keisuke Fujii},
pdfkeywords={First keyword, Second keyword, More},
}

\begin{document}
\maketitle

\begin{abstract}The application of visual tracking to the performance analysis of sports players in dynamic competitions is vital for effective coaching. 
In doubles matches, coordinated positioning is crucial for maintaining control of the court and minimizing opponents' scoring opportunities. The analysis of such teamwork plays a vital role in understanding the dynamics of the game.
However, previous studies have primarily focused on analyzing and assessing singles players without considering occlusion in broadcast videos. These studies have relied on discrete representations, which involve the analysis and representation of specific actions (e.g., strokes) or events that occur during the game while overlooking the meaningful spatial distribution.
In this work, we present the first annotated drone dataset from top and back views in badminton doubles and propose a framework to estimate the control area probability map, which can be used to evaluate teamwork performance. 
We present an efficient framework of deep neural networks that enables the calculation of full probability surfaces.  
This framework utilizes the embedding of a Gaussian mixture map of players' positions and employs graph convolution on their poses. 
In the experiment, we verify our approach by comparing various baselines and discovering the correlations between the score and control area.
Additionally, we propose a practical application for assessing optimal positioning to provide instructions during a game. 
Our approach offers both visual and quantitative evaluations of players' movements, thereby providing valuable insights into doubles teamwork. The dataset and related project code is available at
\url{https://github.com/Ning-D/Drone_BD_ControlArea}
\end{abstract}


\maketitle

\section{Introduction}\label{Introduction}
%general introduction
Visual tracking has become an increasingly popular research area due to its diverse applications in domains such as human-computer interaction, robotics, autonomous driving, and medical imaging \cite{mueller2018ganerated,boutteau2020vision, dasgupta2022spatio, wawrzyniak2019vessel}. Object tracking and pose estimation are two essential components of visual tracking, allowing for accurate and efficient tracking of objects in videos.
%general problem
However, despite significant progress in object tracking \cite{zhang2022bytetrack} and pose estimation \cite{mmpose2020}, applying these technologies to sports fields remains a challenging problem due to partial occlusion, changes in viewpoint, and complex movements of athletes in a dynamic environment.

%broadcast video
As technology has advanced in the past few years, data collection has become more in-depth and can be conducted with relative ease. Significant effort has been focused on building larger broadcast sports video datasets \cite{deliege2021soccernet, giancola2021temporally}. 
%broadcast video problem
However, broadcast videos do not show the entire pitch, only provide partial information about the game, and are mostly available to wealthy professional sports teams. To be specific, there are frequent perspective changes in broadcast video. Besides, broadcast videos also suffer from severe occlusion, especially in team sports. 
In soccer, drone cameras are used to capture the entire pitch in a single frame \cite{scott2022soccertrack}, providing greater adaptability and flexibility. Unfortunately, there is currently no such dataset for any racket sports.

\begin{figure}[ht]
    \centering
    %\centerline{
    \includegraphics[scale=0.43]{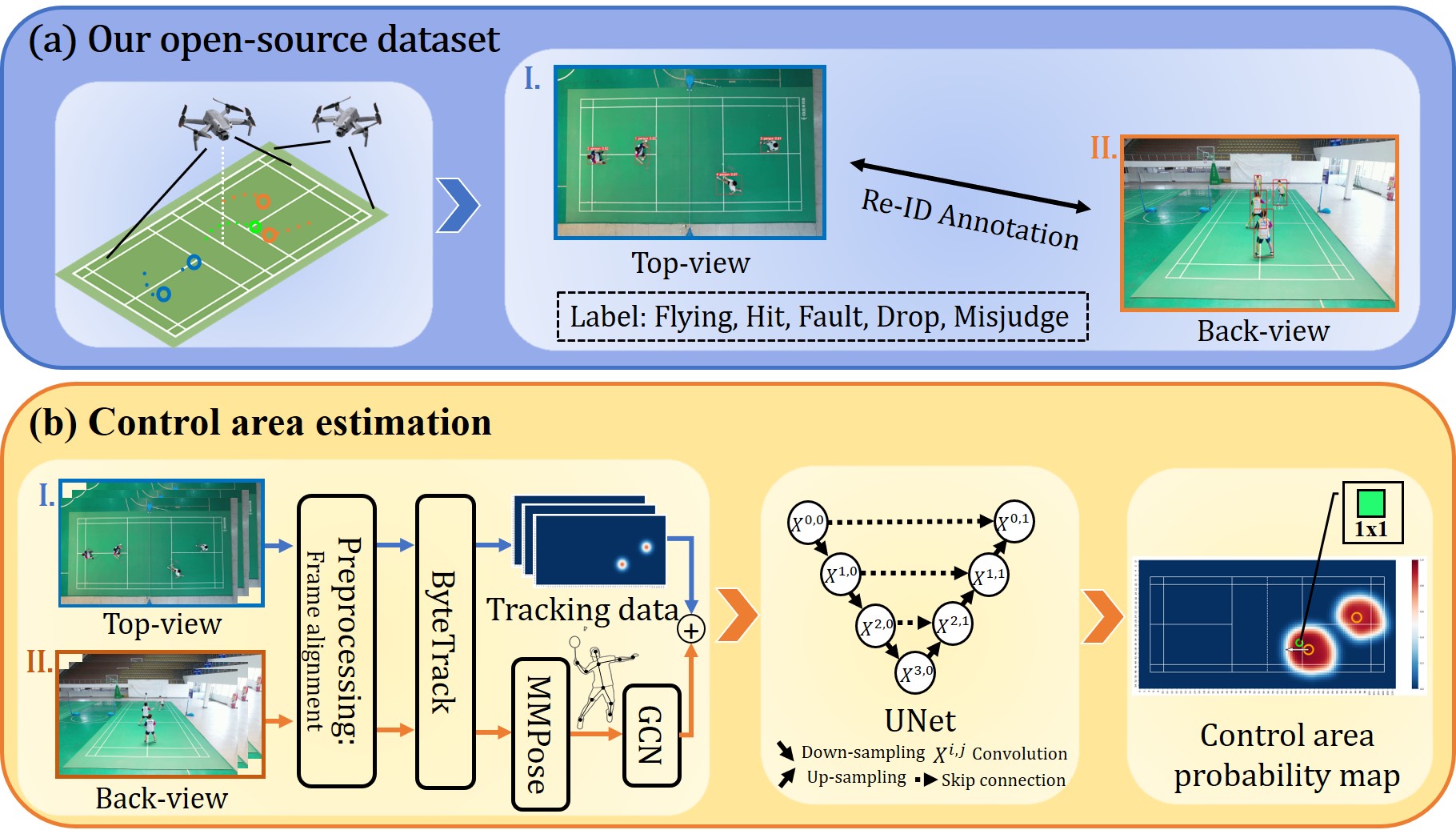}
    \caption{Contribution of our work. (a) Overview of our open-source dataset. (b) Control area estimation as a new visual analysis method.}
    \label{fig:Overview}
\end{figure}

%address method
In this paper, we present a dataset of men's doubles badminton matches, which was captured using two 4K-resolution drone cameras (Fig. \ref{fig:Overview}a). Two drones filmed the court from the top view and back view, respectively, capturing the entire court without being affected by the occlusion problem. Based on this dataset, we propose a novel visual analysis method, as illustrated in Fig. \ref{fig:Overview}b.

%Visual analytics
Visual analytics, including heatmaps, have become prevalent in various sports, especially in team sports like soccer \cite{fernandez2018wide, cho2022pass2vec}. Advanced methods are being developed to estimate probabilities of actions, such as passes, and other performance metrics based on spatiotemporal data. For example, SoccerMap \cite{fernandez2021soccermap} estimates probability surfaces of potential passes from high-frequency spatiotemporal data in team sports like soccer. However, there has been little exploration of fine-grained spatial analysis in racket sports (for more information, please refer to the next section).
%Visual analytics in racket sports
In racket sports, most previous studies have focused on analyzing and assessing singles players in broadcast videos and the discrete representations such as stroke analysis (e.g., \cite{ding2022deep, ghosh2018towards, hsu2019coachai}).
However, these studies tend to overlook meaningful spatial distributions.

Therefore, the objective of this study is to address these limitations by focusing on quantifying spatial value occupation and providing a quantitative position evaluation metric in badminton doubles. To achieve this, we leverage a self-collected drone dataset to estimate the control area probability map, as depicted in Figure \ref{fig:Overview}. Our proposed approach employs a two-stream network architecture (Figure \ref{fig:Overview}b) that combines positional information from a top view with pose information from a back view. By utilizing a 3-layer U-Net model, our method accurately predicts the probability map, allowing for an effective evaluation of the team's control area.

The contributions of this paper are as follows:
\begin{itemize}
\item{We build and share a men’s doubles badminton drone dataset, which includes annotations of bounding boxes, shuttlecock locations (Hit/Drop), and poses. } 
\item{We present an efficient framework of deep neural networks that enables the calculation of full probability surfaces, which utilizes the embedding of a Gaussian mixture map of players' positions and graph convolution of their poses. We validate the effectiveness of this fine-grained analysis of game situations in badminton.}
\item{We also propose a practical application for assessing optimal positioning in badminton that can increase the probability of a successful shot.}
\end{itemize}

This paper is organized as follows. First, in Section \ref{Related work}, We provide an overview of the related work. Next, we describe the badminton drone dataset in Section \ref{Dataset} and explain our approach in Section \ref{Estimations of control area}. Then, we present the experimental results in Section \ref{Results} and conclude this paper in Section \ref{Conclusion}.

\section{Related work}\label{Related work}
\subsection{Visual analytics in team sports}\label{Visual analytics in team sports}

Visual analytics has developed rapidly and is widely used in various sports \cite{article,perin2018state}. In soccer, the Time-to-intercept method is used to calculate a pitch control function that quantifies and visualizes the regions of the pitch controlled by each team \cite{spearman2017physics}. Another approach uses a generative model for multi-agent trajectory data and visualizes the predicted trajectory of players in both soccer and basketball \cite{yeh2019diverse}.

Other game analysis methods use non-sports-specific visualizations. One of these visualizations is the heatmap, which visualizes the most frequent locations of game events by density. 
%Recent work has proposed a deep learning architecture to estimate the full probability surfaces of potential passes in soccer \cite{fernandez2021soccermap}. 
In the study conducted by Fernandez et al. \cite{fernandez2021soccermap}, a deep learning architecture was proposed to estimate the probability surfaces of potential passes in soccer, which is closely related to our research. However, a crucial aspect that was overlooked in their study is the impact of player poses on the probability maps. 
SnapShot \cite{pileggi2012snapshot} introduced a specific type of heatmap called radial heatmap to display shot data in ice hockey, while CourtVision \cite{goldsberry2012courtvision} quantifies and visualizes the shooting range of players in basketball. Another commonly used visualization method is the flow graph, where the nodes' size shows each player's role, and the links show the connections between them \cite{perin2013soccerstories}. Besides, the glyph-based visualization method has also been applied to sports. For example, MatchPad \cite{legg2012matchpad} adopts a glyph-based visual design to analyze the performances of players during rugby games. All these works demonstrate the need, impact, and potential of visual analytics in sports.

\subsection{Visual analytics in racket sports}\label{Visual analytics in racket sports}
In racket sports, most previous studies focused on analyzing and assessing singles players and visualizing discrete representations (e.g., stroke). In tennis, CourtTime \cite{polk2019courttime} introduced a novel visual metaphor to facilitate pattern detection, while Tennivis \cite{polk2014tennivis} visualized statistical data, such as score and service information. In table tennis, Tac-miner \cite{wang2021tac} developed a visual analytics system to facilitate simulative analysis based on the Markov Chain, while iTTVis \cite{wu2017ittvis} used a matrix to reveal the relationship among multiple attributes within strokes. Tac-Simur \cite{wang2019tac} provided an interactive exploration of diverse tactical simulation tasks and visually explained the simulation results. In badminton, TIVEE \cite{chu2021tivee} studied tactic analysis in a 3D environment and proposed immersive visual analytics.
Recently, Haq et al. \cite{haq2022heatmap} utilized player tracking to visualize position statistics using heatmaps. However, this approach suffers from drawbacks such as information loss during data aggregation and reliance on implicit assumptions about data distribution. In contrast, our approach offers higher precision, detailed representations, and the capability to handle dynamic data for temporal analysis.
%However, exploiting spatial relationships and accurately estimating observed and unobserved game events on the field in racket sports remains an area that has received little exploration.

\subsection{Racket sports datasets}\label{Racket sports datasets}
Broadcast videos have been widely used as the dataset in racket sports. Recently, significant efforts have been made to build larger broadcast sports video datasets. In badminton, YouTube videos from the Badminton World Federation have been used in academic data analysis studies \cite{archana2016efficient, ding2022deep}. However, broadcast videos often suffer from frequent perspective changes and occlusion issues \cite{archana2016efficient, ghosh2018towards}. 

Other works have collected and built their dataset based on specific task requirements, providing an optimal perspective for analysis. For instance, in tennis, TTNet \cite{voeikov2020ttnet} introduced the OpenTTGames dataset for game events detection and semantic segmentation tasks. In table tennis, Blank et al. \cite{blank2015sensor} attached inertial sensors to rackets to collect stroke data, while Kulkarni et al. \cite{kulkarni2021table} positioned their cameras and vibration sensors to capture the most optimum view for detailed stroke analysis.

\section{Dataset}\label{Dataset}
\subsection{Video collection}\label{Video collection}

We collected our data from 2-vs-2 men's doubles badminton games played among members of a college badminton club. Prior to data collection, we obtained approval from Anhui Normal University's ethics committee (approval number [AHNU-ET2022042]) on 14th April 2022, and we conducted the study in compliance with the principles of the Declaration of Helsinki. All participants provided signed informed consent.
To capture the entire badminton court, we used two DJI Air 2S drones (Da-Jiang Innovations Science and Technology Co., Ltd., China) that provided top and back views. The video resolution was 4K (3,840 $\times$ 2,160 pixels), and the frame rate was 30 fps. Our raw video data included 39 games, involving 14 pairs, 11 players, and a total of 1347 rallies.

It is important to highlight that publicly available badminton datasets primarily consist of broadcast videos sourced from the Badminton World Federation (BWF) channel on YouTube. However, these videos typically only offer back-view footage and lack the crucial top-view videos required for our proposed method.

\subsection{Data annotation and structure}\label{subsec2}
This section outlines our approach for efficiently annotating bounding boxes and shuttlecock locations in drone videos (Fig. \ref{fig:Overview}a). Manual annotation is a time-consuming process that can take several hours to annotate a single rally. Therefore, we utilized well-established computer vision techniques to shorten the process as shown in Fig. \ref{fig:fig/data_flow}.

The dataset is structured such that each rally is accompanied by two annotation files in a simple comma-separated value (CSV) format. For shuttlecock detection, each line of the CSV file contains five values: frame number, visibility, x-coordinate, y-coordinate, and status. The status of the shuttlecock in each frame was annotated with one of five types: Frying, Hit, Fault, Drop, and Misjudge. For players, we also provide corresponding bounding box coordinates in CSV format. 

To track players and the shuttlecock in the raw video data, we first use homography transformation to eliminate the offset problem caused by the drone's perspective. We segmented each game into several rallies, and for each rally, we estimated the XY-coordinate values for the locations of the four players using tracking, specifically ByteTrack \cite{zhang2022bytetrack}, a popular high-precision multi-object detection and tracking system. We detected the shuttlecock location using TrackNet \cite{huang2019tracknet}, an object tracking network that has been proven to exhibit decent tracking capability in games that involve small, high-speed balls such as shuttlecocks.
We adjusted any outliers using a simple labeling tool with a MATLAB GUI, modifying the code originally created and utilized in the study of TrackeNet \cite{huang2019tracknet} to fit our study's requirements.
%re_ID annotation
Moreover, we utilized direct linear transformation (DLT) to re-identify players' IDs and manually corrected any discrepancies.
By combining detection results with manual adjustments, we were able to significantly expedite the annotation process.

\begin{figure}[ht]
    \centering
    \includegraphics[scale=0.6]{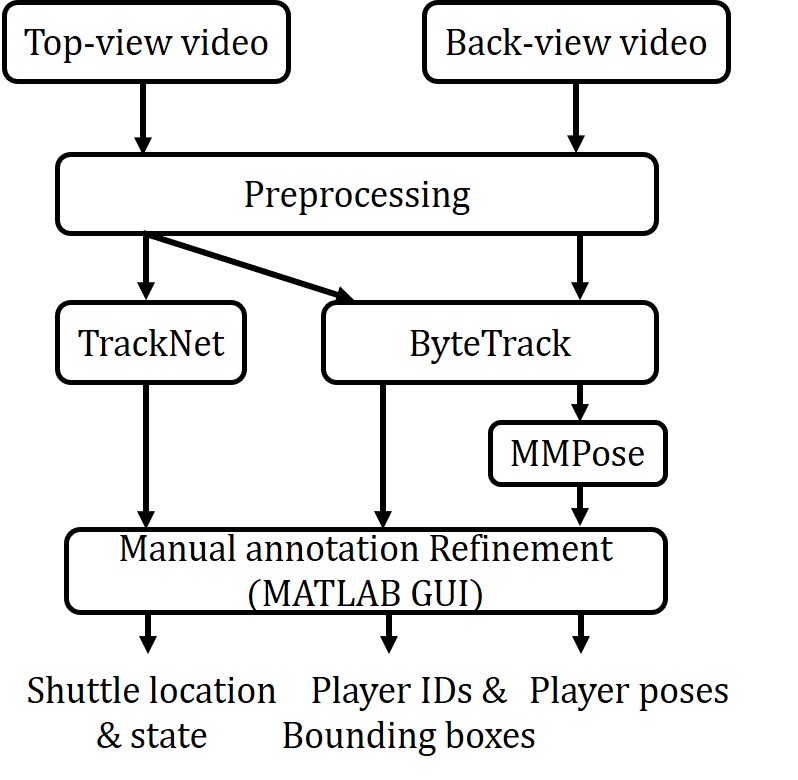}
    \caption{Dataset annotation workflow for top-view and back-view drone videos.}
    \label{fig:data_flow}
\end{figure}

\section{Estimations of control area}\label{Estimations of control area}
We propose a framework for estimating the probability map of the control area. To achieve this, we construct a neural network that learns the relationship between tracking/event data and the map in a data-driven manner. In this section, we will describe the architecture of the neural network and the learning method.

\begin{figure}[ht]
    \centering
    \includegraphics[scale=0.4]{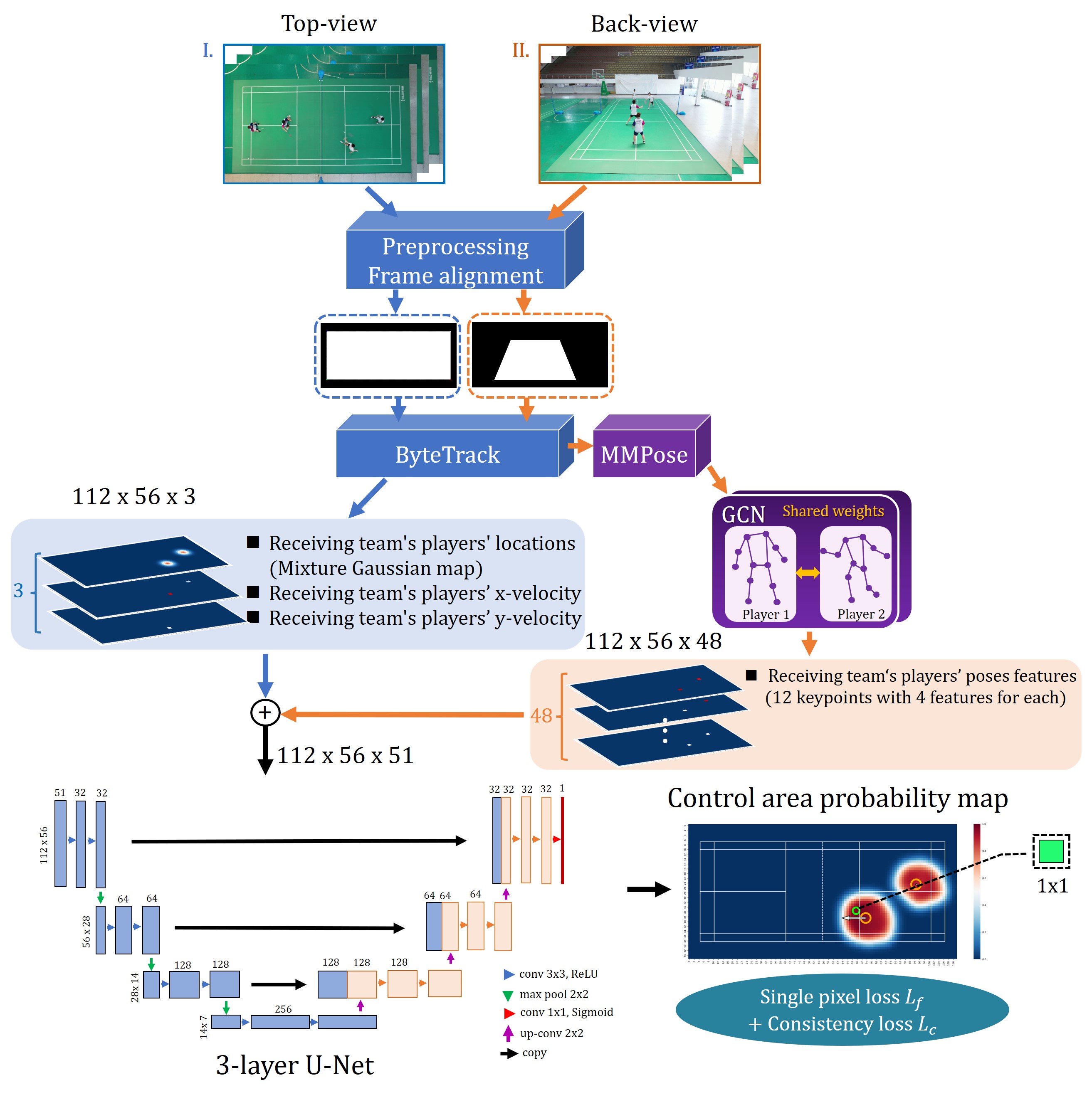}
    \caption{Detailed architecture diagram of the proposed model, showing the two-stream network and the 3-layer U-Net model used to generate the control area probability map.}
    \label{fig:Maps}
\end{figure}

\subsection{Model architecture}\label{Model architecture}

The model architecture consists of a two-stream network. The first stream is based on the top view and captures information about the location and velocity of players, while the second stream is based on the back view and captures information about the pose of players. The outputs of the two streams are then combined to serve as the input of a 3-layer U-Net model, which predicts the control probability map.

A 3-layer U-Net network was constructed to estimate the full probability map of the control area, considering that the input image size is 112 $\times$ 56. Here, the term `3-layer' refers to the fact that both the downsampling layer (Max pooling) and the upsampling layer (Up-convolution) in the network consist of three layers each, as shown in Figure \ref{fig:Maps}. U-Net \cite{ronneberger2015u} is a convolutional network architecture specifically designed for fast and accurate image segmentation. We expect that this network can be effectively utilized in our work to segment and identify areas that are controllable or not.
The network takes as input: (1) a Gaussian mixture probability map centered on the location of 2 players (same team) who receive the shuttlecock, (2) the X-velocity and Y-velocity of 2 players (same team) who receive the shuttlecock, and (3) the poses of 2 players (same team) who receive the shuttlecock. The location where the player hits the shuttlecock or where the shuttlecock lands (drop) is used as the target location to obtain the control area probability map.

\subsection{Learning}\label{Learning}
Our proposed loss function $L$ consists of a Focal loss and a constraint on spatial continuity, denoted as follows:

\begin{equation}
L=L_{f}+\mu L_c
\end{equation}

\noindent Here, $\mu$ represents the weight for balancing the two constraints, where we set $\mu=0.03$. As the objective function of our model, we use Focal Loss \cite{lin2017focal} to address the class imbalance problem in our dataset, defined as:

\begin{equation}
p_{\mathrm{t}}= \begin{cases}p & \text { if } y=1 \\ 1-p & \text { otherwise, }\end{cases}
\end{equation}

\begin{equation}
F L\left(p_{\mathrm{t}}\right)=-\alpha \left(1-p_{\mathrm{t}}\right)^\gamma \log \left(p_{\mathrm{t}}\right),
\end{equation}

\noindent where $p_t$ is the estimated probability of the model. We set $\alpha=0.8$ and $\gamma=3$.
The focal loss $L_{f}$ can then be written as:

\begin{equation}
L_{f}=F L\left(y_{l o c_k}, f\left(x_k ; \theta\right)_{l o c_k}\right),
\end{equation}

\noindent where $x_{k}$ is the game state at time $k$, $loc_{k}$ represents the location where the player hit the shuttlecock or where the shuttlecock dropped at time $k$, and $y_{lock}$ represents the ground truth control probability at time $k$. 

We also use an additional constraint $L_{c}$, similar to Kim et al. \cite{kim2020unsupervised}. This loss term is used to encourage spatial smoothness and continuity in the estimated control probabilities.
The spatial continuity loss is defined as follows:

\begin{equation}
L_c=\sum_{i=1}^{W-1} \sum_{j=1}^{H-1}\left\|v_{i+1, j}-v_{i, j}\right\|_1+\left\|v_{i, j+1}-v_{i, j}\right\|_1,
\end{equation}

\noindent where $W$ and $H$ represent the width and height of the image, respectively. $v_{i,j}$ denotes the control probability of the pixel at coordinates $(i, j)$.
The loss is calculated by summing the $L_1$-norm differences between adjacent pixels in both horizontal and vertical directions. 
Minimizing this loss term encourages the network to generate spatially continuous control probabilities with smooth transitions and discourages the presence of complex patterns or abrupt changes in the probability map. At the same time, it can also mitigate the impact of data amount on the results to some extent.

The network is trained using the Adam optimizer with a learning rate of $10^{-6}$, 30 epochs, and batch sizes of 16. For the learning, we augmented our dataset with a horizontal flip. We have 12,658 hit samples and 796 drop samples. For the training phase, we used a ratio of 0.8 hit samples and 0.5 drop samples, and the rest for testing.

\subsection{Optimal positioning}\label{Optimal positioning}

Our model can estimate the control area and evaluate the players in badminton doubles. However, determining the optimal positioning and movement strategy for players, which is a critical aspect of successful performance, is currently unknown. In this study, we propose an approach to identify optimal positioning strategies for doubles badminton players based on data-driven analysis.

\begin{figure}[ht]
    \centering
    \includegraphics[scale=0.25]{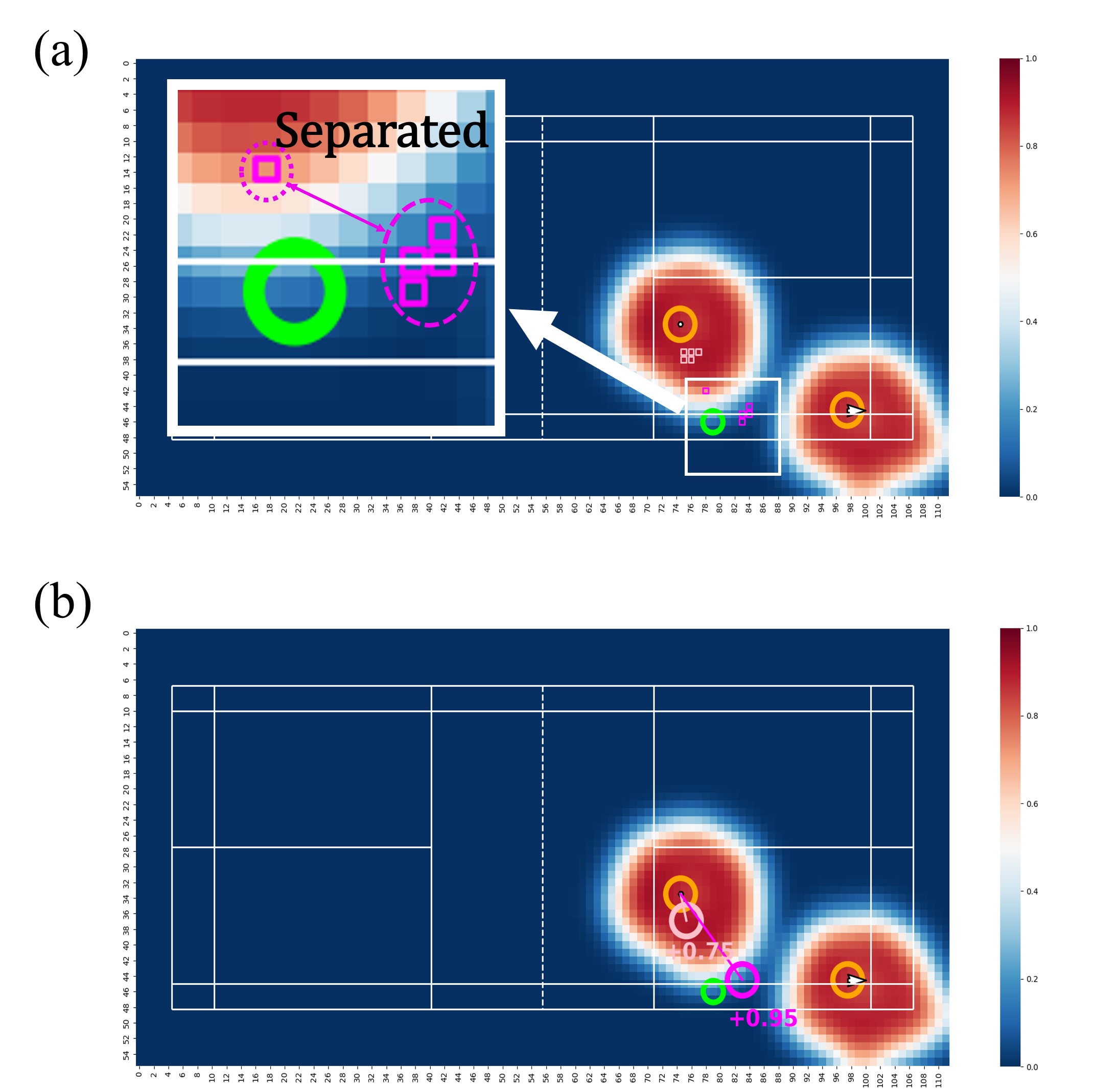}
    \caption{(a) This is an example of five optimal position candidates (magenta rectangles) within the control probability of $P_c(x,y) >= 0.95$, which are separated into two clusters, illustrating the need for unsupervised clustering. (b) The recommended optimal position in magenta circle is the average value of all locations in the largest cluster.}
    \label{fig:Separated}
\end{figure}

We define the control probability function $P_c(x,y)$ as a function that measures the probability of successfully controlling the shuttlecock when the receiver is located at grid location $(x,y)$, and the teammate is located at the same position as in the actual play. The notation $P_c(x,y) >= p$ denotes the grid location where the control probability is greater than or equal to $p$.
To reduce the impact of randomness or variability, we first select the set of $n$ nearest grid locations to the actual play position of the receiver that are within $P_c(x,y) >= p$. However, we note that these $n$ nearest grid locations may be separated and not closely located together as shown in Fig. \ref{fig:Separated}a.
To determine the optimal location, we use unsupervised clustering, specifically the hierarchy clustering \cite{johnson1967hierarchical}, to group these $n$ locations into clusters and identify the largest cluster $C_{max}$ among them. We then calculate the average value of all locations in $C_{max}$ to obtain the recommended position for the receiving player that increases the probability of controlling the shuttlecock to $p$ as shown in Fig. \ref{fig:Separated}b. Eq. (\ref{eq:optimal}) shows that the recommended position for the receiver is obtained by calculating the average value of all grid locations in the largest cluster $C_{max}$, which is denoted by $(x_{rec}, y_{rec})$. The formula is defined as:

\begin{equation}\label{eq:optimal}
\left(x_{r e c}, y_{r e c}\right)=\frac{1}{ n (C_{\max })} \sum_{(x, y) \in C_{\max}}(x, y)
\end{equation}

\noindent Here, $n(C_{max})$ denotes the number of grid locations in the largest cluster. Overall, our approach provides a data-driven and actionable recommendation for players and coaches to follow in practice and competition.

\section{Results}\label{Results}

In this section, we first verified the control area estimation model by visualizing the estimated control area and evaluating the accuracy. Second, we examined the practical usefulness of our approach by investigating the relationship between the control area and the score.

\begin{figure}[ht]
    \centering
    \includegraphics[scale=0.7]{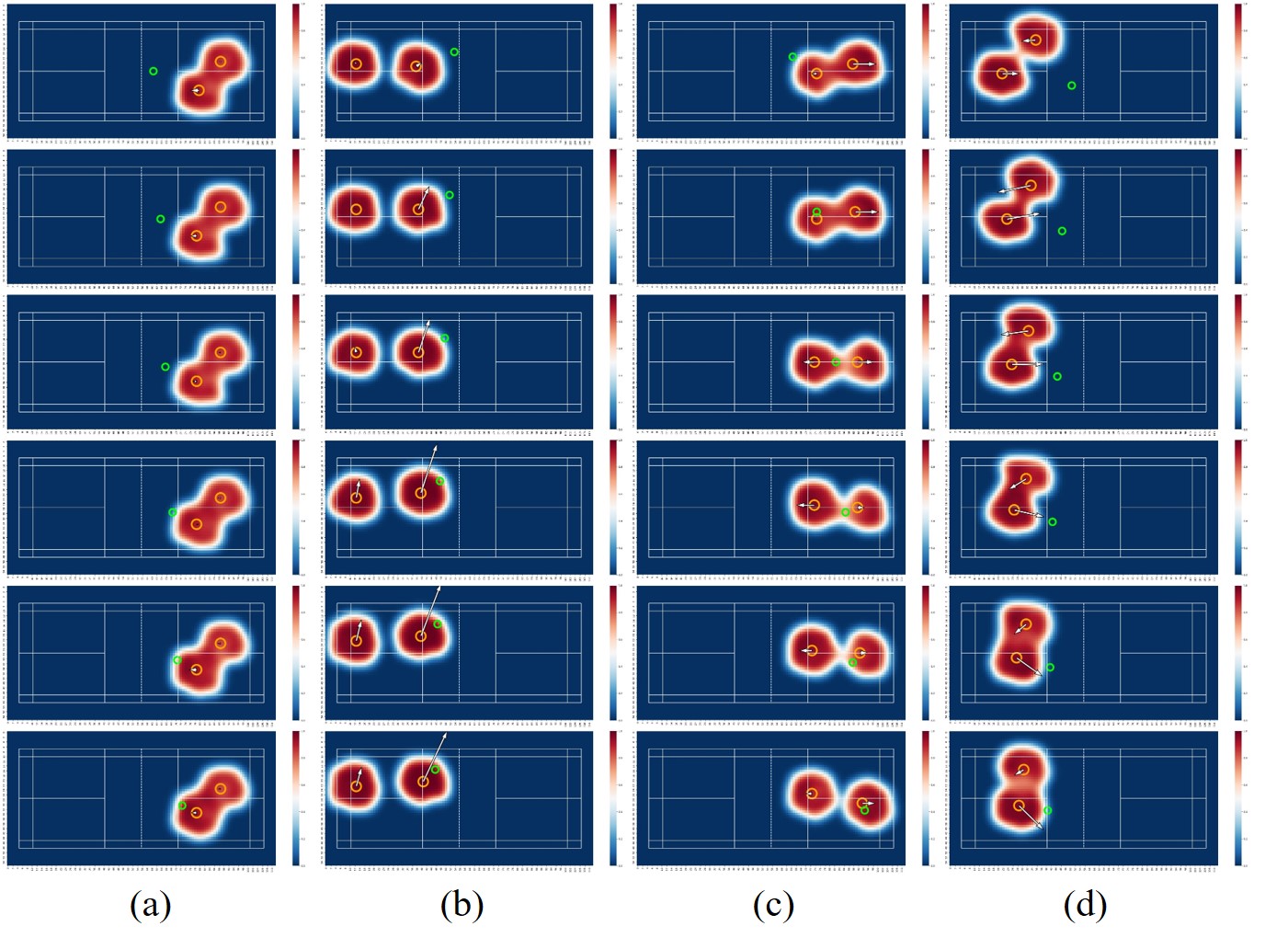}
    \caption{Visualization of control area changes of the receivers during catches in a rally (time passes from top to bottom). Orange and lime circles represent players’ locations and the shuttlecock location, respectively, while the arrows indicate the velocity vector for each player. A darker shade of red denotes a higher probability of control.}
    \label{fig:CA}
\end{figure}
\vspace{-5pt}

\subsection{Control area estimation}\label{Control area estimation}
First, we visualized the control area when a team reacts to an incoming shuttlecock after the shuttlecock crosses the net. The changes in the control area can help to understand the reasons for losing points.  
Fig. \ref{fig:CA} presents the changes in the control area probability map of the receivers (on both sides) during a catch in a rally. Receivers can hit (receive) the shuttlecock in cases shown in Fig. \ref{fig:CA}a, b, and c. In the last case shown in Fig. \ref{fig:CA}d, the left-side receivers failed to catch the shuttlecock. We did not make any assumptions about the shape of the distribution in the control area (i.e., learned from data). Additionally, we observed that the model may learn the player’s speed for estimating the occupied spaces of the control area.

\subsection{Verification of our method}\label{Verification of our method}

In this study, we evaluated the performance of our model and examined the impact of players' velocity ($-$ Players' velocity), players' pose  ($-$ Pose), and computed bounding box (Bbox) height and width from top-view ($-$ Pose $+$ Bbox (top)) on the estimation performance. 

The last one replaced the players' poses in the full model with height and width measurements of bounding boxes (bboxes) as input to examine the effectiveness of the pose information. 
To evaluate the model, we computed the $L_1$ classification loss between the ground truth and estimated positions for hit/drop samples. As presented in Table \ref{tab:table1}, a model trained with the full components (players' velocity and pose) achieved the best performance for both control (hit) and non-control (drop) samples, indicating that both the players' velocity and pose information contributed to accurately estimating the control area probability map. Notably, Bbox (top) was not included in the full components.
%-Pose+BBox (top)
We also found that using the back view poses produced more accurate results compared to using the bboxes from the top view.
In the model verification stage, we achieved an overall $L_1$ classification loss of 0.094 for the test samples with hit and drop shuttlecocks. Specifically, the $L_1$ classification loss for hit and drop samples was 0.085 and 0.238, respectively.

\begin{table*}[ht]
    \begin{center}
        \caption{Comparison of $L_1$ classification loss for each input feature by eliminating the different input features from the total input.}
            
            \begin{tabular}{c c c c c}
            \hline
            Error & $-$Players' velocity & $-$Pose & $-$Pose $+$ Bbox (top) & Full \\ 
            \hline
            Control (hit)  & 0.118 &   0.100 & 0.092 &  \textbf{0.085} \\ 
            Non-control (drop)  & 0.244 &  0.294 & 0.252 & \textbf{0.238} \\ 
            All  & 0.125  & 0.110 & 0.101 & \textbf{0.094} \\ 
            \hline
            \end{tabular}
            
    \label{tab:table1}
    \vspace{0pt}
    \end{center}
\end{table*}

\subsection{Relationship with the score}\label{Relationship with the score}

\subsubsection{Control area in the full field}\label{Control area in the full field}
\begin{figure}[ht]
    \centering
    %\centerline{
    \includegraphics[scale=0.6]{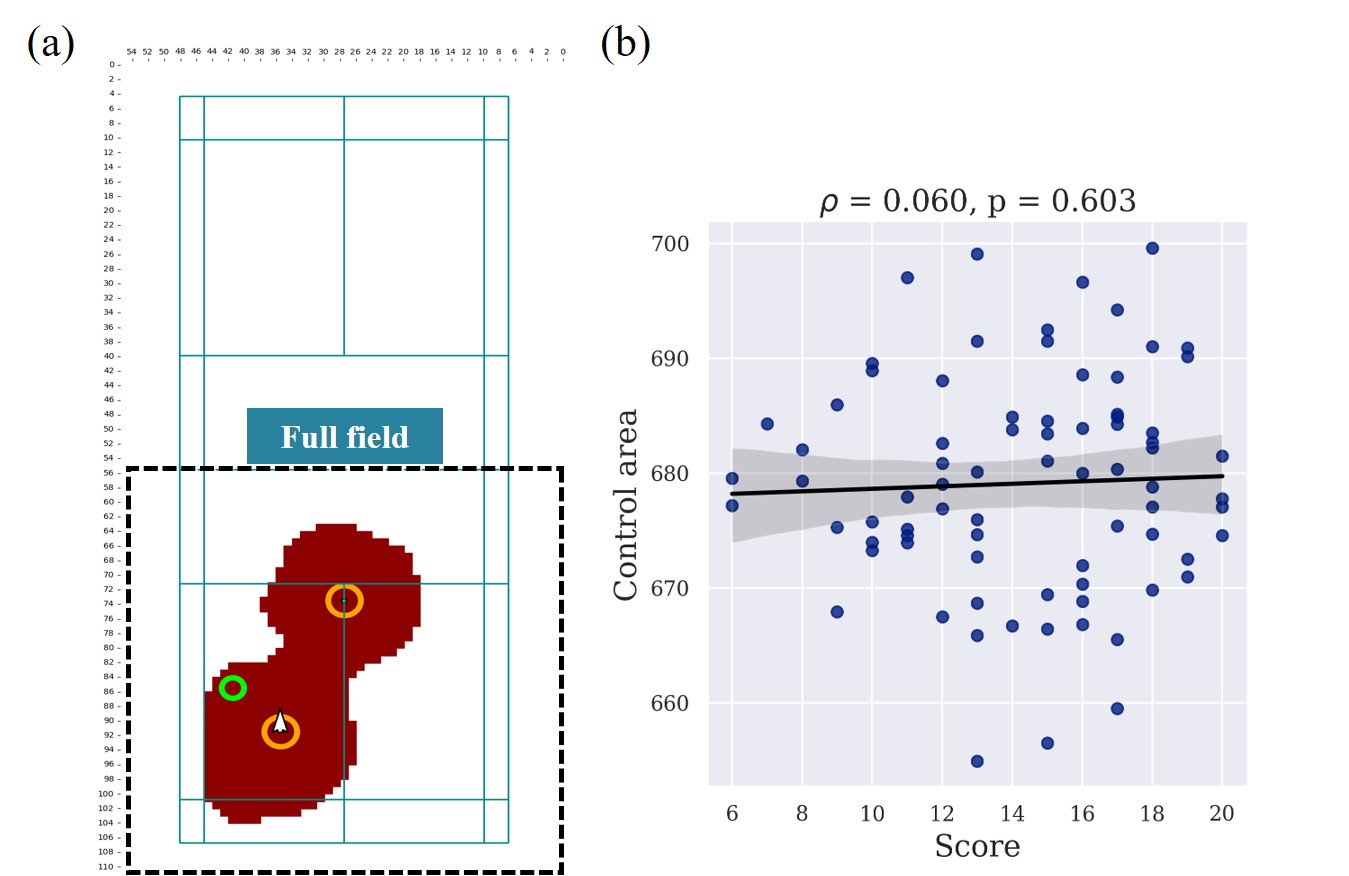}
    \caption{(a) Definition of the full field. (b) Correlation between the score and the size of the control area.}
    \label{fig:Full}
\end{figure}

We defined the full field as half of the input map size ($56 \times 56$) on the side of the receiving team, as shown in Fig. \ref{fig:Full}a. First, we examined the relationship between the score and the size of the control area on the full field. Fig. \ref{fig:Full}b indicates that there is no correlation between the score and the size of the control area ($p > 0.05$). We speculate that the size of the control area may be related to the velocity of two players rather than the team’s performance (defense capabilities). Therefore, in the next section, we analyzed the size of the control area in the primary area instead of the full field.

\vspace{-5pt}

\begin{figure}[ht]
    \centering
    %\centerline{
    \includegraphics[scale=0.5]{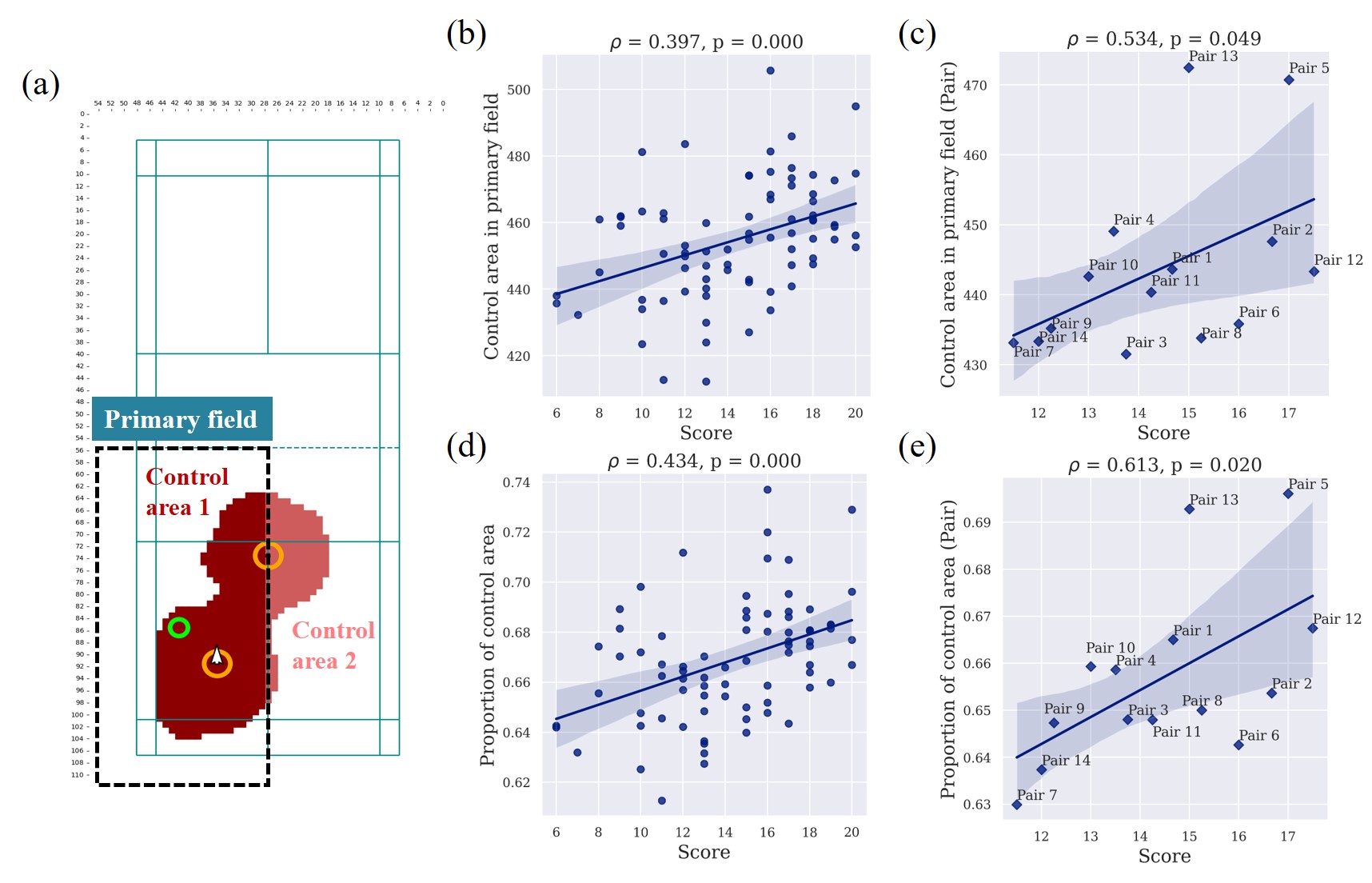}
    \caption{(a) An example of how we define the primary field when the shuttlecock is located at the green circle is as follows:
the primary field is the rectangular area that includes the green circle and covers one-quarter of the input map size ($56\times 28$).
    (b-e) Correlations between the score and the (b,c) control area in the primary field or (d,e) proportion of the control area for each (b,d) game or identified pair (c,e)}. 
    \label{fig:Primary field}
\end{figure}

\subsubsection{Control area in primary field}\label{Control area in primary field}

Next, we defined the primary field as the field where the shuttlecock is located, which has a size of one-quarter of the input map size ($56\times 28$).
In Fig. \ref{fig:Primary field}, we analyzed the control area in the primary field, which depends on the shuttlecock location, and the proportion of the control area to evaluate the team's performance in coverage, i.e., defense capabilities. The score indicates the number of points scored by the team in a game, and the control area in the primary field refers to $C_{\text {primary}}=$ control area 1 if the shuttlecock is located in the left of the field as shown in \ref{fig:Primary field}a. The proportion of the control area refers to $P_{\text {control area }}=\frac{\text { control area } 1}{\text { control area } 1+\text { control area } 2}$ if the shuttlecock is located in the left of the field.

\begin{figure}[h]
    \centering
    \includegraphics[scale=0.5]{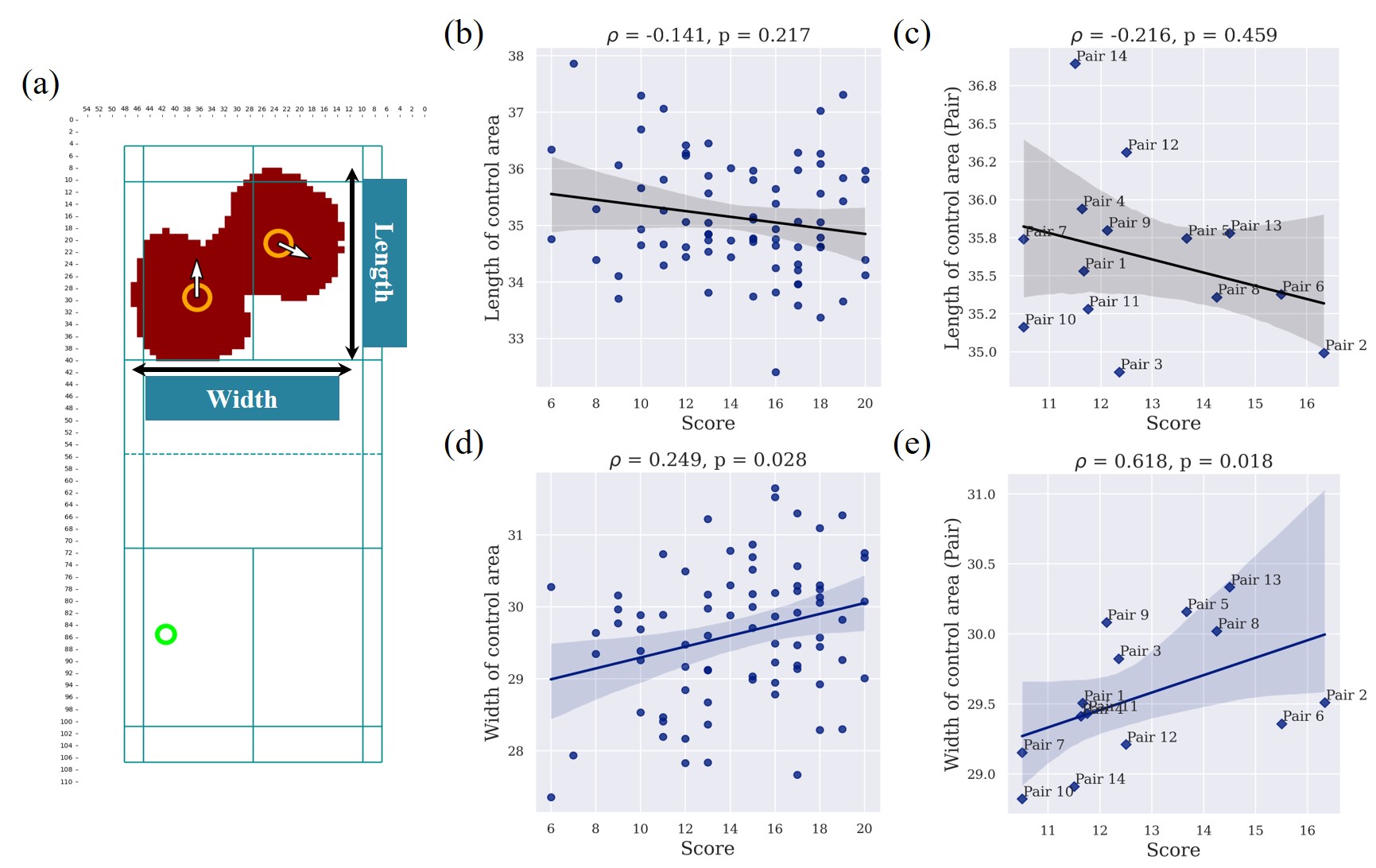}
    \caption{(a) Definition of the length and width of the control area of a team. (b-e) Correlations between the score and the (b,c) length or (d,e) width of the control area for each (b,d) game or identified pair (c,e). 
    }
    \label{fig:LW}
\end{figure}

We found moderate positive monotonic correlations between the score and the control area in the primary field (Fig. \ref{fig:Primary field}b, $\rho = 0.397$ ($p < 0.05$)) as well as between the proportion of the control area (Fig. \ref{fig:Primary field}d, $\rho = 0.434$ ($p < 0.05$)).
In addition, for each pair, we observed a strong positive monotonic correlation between the score and the control area in the primary field (Fig. \ref{fig:Primary field}c, $\rho = 0.534$ ($p < 0.05$)), as well as between the proportion of the control area (Fig. \ref{fig:Primary field}e, $\rho = 0.613$ ($p < 0.05$)). These results suggest that the pair with better team performance tends to have a larger control area in the primary field where the shuttlecock is located.

\subsubsection{Length/width of the control area}\label{Length/width of the control area}
In badminton, mastering the ``doubles rotation'' skill is crucial to maintain court coverage and preventing any gaps during the match. The ability to effectively cover the field is a valuable indicator of player performance.

To measure field coverage, we analyzed the control area of each team at the moment their opponents hit the shuttlecock. This is the time when both players in the team should prepare and position themselves for the next stroke (see Fig. \ref{fig:LW}a).

For each game and each pair, we found no correlation between the score and the length of the control area ($p > 0.05$), as shown in Fig. \ref{fig:LW}b and c. However, we observed a weak positive monotonic correlation between the score and the width of the control area for each game (Fig. \ref{fig:LW}d, $\rho = 0.249$ and $p < 0.05$), and a strong positive monotonic correlation between the score and the width of the control area for each pair (Fig. \ref{fig:LW}e, $\rho = 0.618$ and $p < 0.05$). These results suggest that teams with better performance tend to cover more width of the field when preparing for the next stroke.

\subsubsection{Aiming technique}\label{Distance}

\begin{figure}[ht]
    \centering
    %\centerline{
    \includegraphics[scale=0.5]{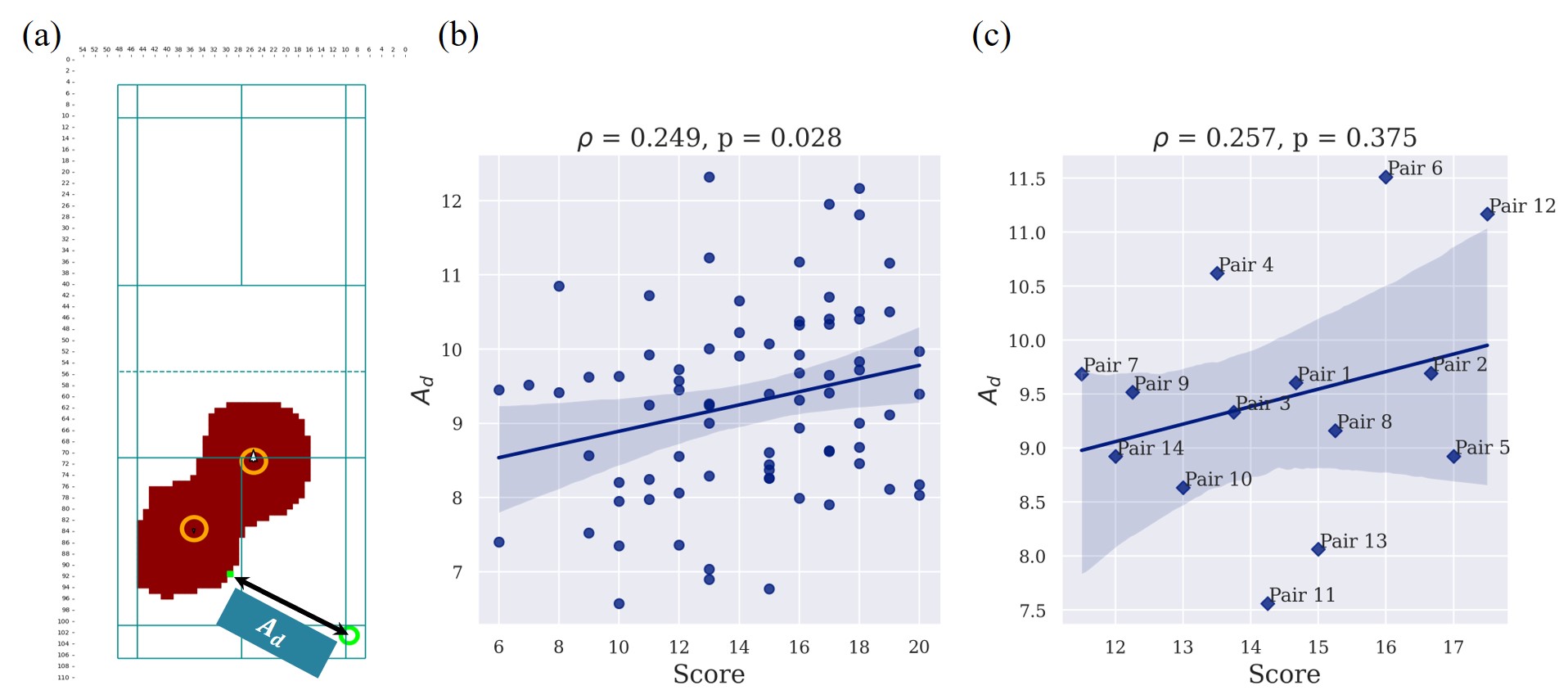}
    \caption{(a) Definition of the aiming technique distance $A_{d}$. (b) Correlation between the score and $A_{d}$ for each game. (c) Correlation between the score and $A_{d}$ for identified pair.}
    \label{fig:Aim distance}
\end{figure}

In doubles badminton, players look for opportunities to hit the shuttlecock towards areas where the opposing team's formation is relatively distant during the moment of hitting. This strategy aims to increase the likelihood of their opponents making an error.
Therefore, we use the maximum distance between the position where the player aims to land the shuttlecock and the control area of opposing players across all rallies as a measure of this aiming technique, denoted as $A_{d}$, as shown in Fig. \ref{fig:Aim distance}a.  In cases where the player's shot is returned by the opponent, we approximate the player's aiming position with the opponent's actual hitting position.

We observed a weak positive monotonic correlation between the score and the $A_{d}$ for each game (Fig. \ref{fig:Aim distance}b, $\rho = 0.249$ and $p < 0.05$)
For each pair, we found no correlation between the score and the  $A_{d}$ ($p > 0.05$), as shown in Fig. \ref{fig:Aim distance}c. 
These results suggest that the aiming technique measured by $A_{d}$ has some impact on the overall score in doubles badminton games. However, for individual pairs, other factors such as teamwork, communication, and individual skills, may also play important roles in determining the success of a doubles badminton pair.

%\section{Practical applications}\label{Practical applications}
\subsection{Assessment of optimal positioning}\label{Assessment of optimal positioning}

As proposed in Section \ref{Optimal positioning}, we define optimal locations as those that are near the receiver and provide a higher probability of a successful shot than the current location. To improve the chances of controlling the shuttlecock during drop shots in doubles, we recommend moving along the shortest path while maintaining the hitting pose. Specifically, we suggest fixing one player's location and considering all possible grid locations ($112 \times 56$) for the receiving player. 

\begin{figure}[ht]
    \centering
    %\centerline{
    \includegraphics[scale=0.45]{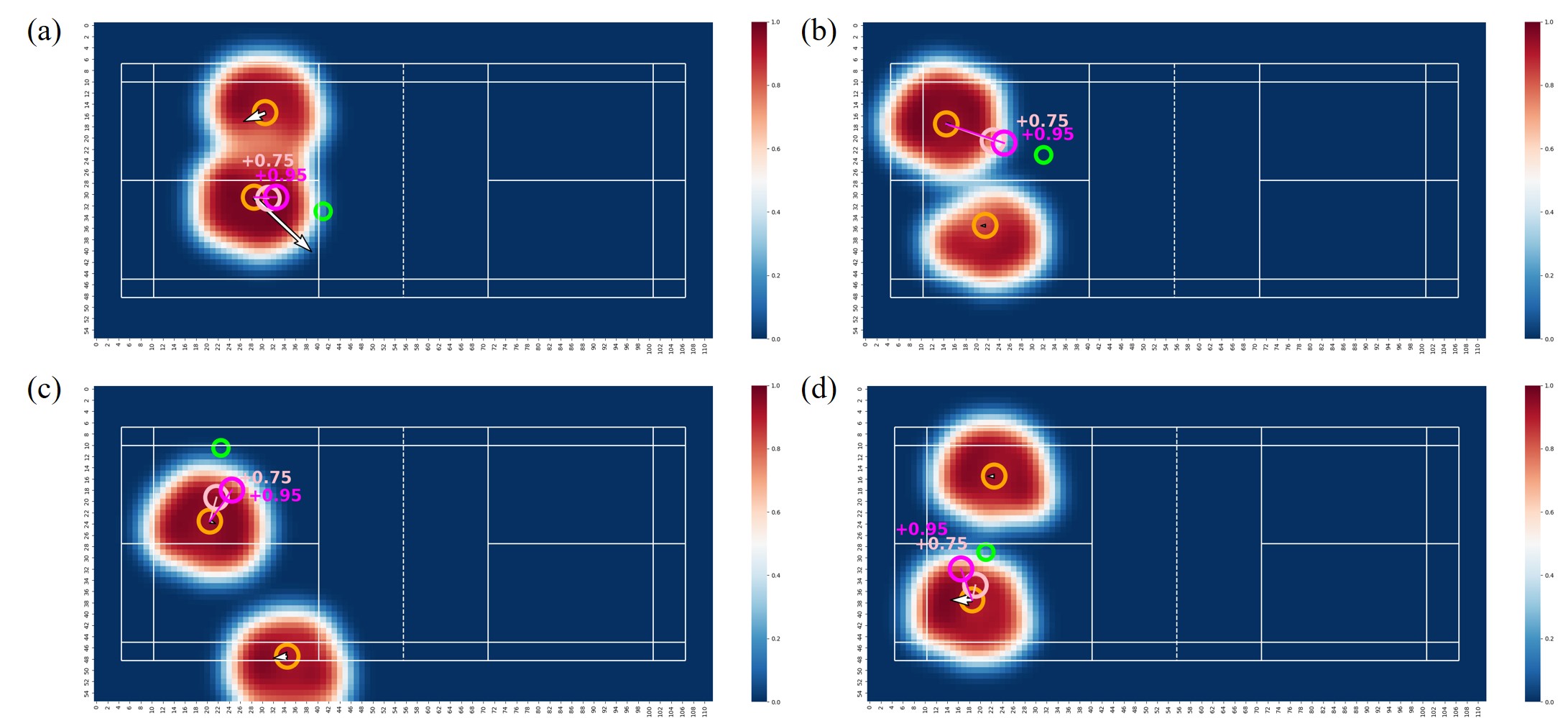}
    \caption{(a-d) The figures present the optimal positions for drop samples. The orange circles indicate the current positions of both players, while the plum and magenta circles represent the recommended positions for the receiver. By positioning the receiver in these locations, the probability of successfully controlling the shuttlecock can increase to either 0.75 or 0.95.}
    \label{fig:Optimal}
\end{figure}

In our case, we use our control probability model to identify the five nearest grid locations ($n=5$) to the actual play position of the receiver where the probability of controlling the shuttlecock is greater than or equal to 0.75 and 0.95 ($P_c(x,y) >= 0.75$ and $P_c(x,y) >= 0.95$).
We then use a hierarchical clustering algorithm to group these five locations into clusters and identify the largest cluster among them.
Finally, we calculate the average value of all locations in the largest cluster, which gives us a recommended position for the receiving player that increases the probability of controlling the shuttlecock to 0.75 (plum circle) and 0.95 (magenta circle), as shown in Fig. \ref{fig:Optimal}.

\section{Conclusion}\label{Conclusion}

In this study, we developed a framework to estimate the control area probability map from an input badminton drone video. We verified our approach by comparing it with various baselines and discovered valuable insights into the relationship between the score and control area. We also shared the first annotated badminton drone video dataset and provided a practical solution for evaluating optimal positioning in badminton, which can improve the likelihood of successful shots. We believe this visual tool can be extended to other racket sports. Our approach can evaluate players' movements both visually and quantitatively, providing valuable insights into doubles teamwork for coaching, assessing, and scouting purposes.

In our future work, we plan to consider more dynamic indicators to reflect the skill in \ref{fig:Aim distance}a, and extend our framework to a variety of other racket sports, such as table tennis and tennis. We believe such visual representation of complex information can provide coaches with a deep perception of the game situation, thus providing a competitive advantage to an individual or a team.

\bibliographystyle{unsrtnat}
\bibliography{reference}

\end{document}